\documentclass[10pt,twocolumn,letterpaper]{article}

\usepackage{iccv}
\usepackage{times}
\usepackage{epsfig}
\usepackage{graphicx}
\usepackage{amsmath}
\usepackage{amssymb}
\usepackage{array}
\usepackage{booktabs}
\usepackage{pifont}
\usepackage[inline]{enumitem}

\usepackage[pagebackref=true,breaklinks=true,letterpaper=true,colorlinks,bookmarks=false]{hyperref}

\iccvfinalcopy 

\def\iccvPaperID{588} 

\usepackage[square,numbers,sort&compress]{natbib}
\usepackage{enumitem}
\usepackage{algpseudocode}
\usepackage{multirow}
\usepackage{booktabs}
\usepackage{algorithm}
\usepackage{caption}
\usepackage{subcaption}
\usepackage[normalem]{ulem}
\usepackage{xparse}

\newcommand{\bfline}[1]{\textbf{\underline{#1}}}

\ificcvfinal\fi
\begin{document}

\title{Cut, Paste and Learn: Surprisingly Easy Synthesis for Instance Detection}

\author{
\large Debidatta Dwibedi \quad\quad Ishan Misra \quad\quad Martial Hebert\\
\large The Robotics Institute, Carnegie Mellon University\\
{\tt\small debidatta@cmu.edu, \{imisra, hebert\}@cs.cmu.edu }} 
\maketitle

\begin{abstract}
A major impediment in rapidly deploying object detection models for instance detection is the lack of large annotated datasets. For example, finding a large labeled dataset containing instances in a particular kitchen is unlikely. Each new environment with new instances requires expensive data collection and annotation. In this paper, we propose a simple approach to generate large annotated instance datasets with minimal effort. Our key insight is that ensuring only patch-level realism provides enough training signal for current object detector models. We automatically `cut' object instances and `paste' them on random backgrounds. A naive way to do this results in pixel artifacts which result in poor performance for trained models. We show how to make detectors ignore these artifacts during training and generate data that gives competitive performance on real data. Our method outperforms existing synthesis approaches and when combined with real images improves relative performance by more than $21\%$ on benchmark datasets. In a cross-domain setting, our synthetic data combined with just $10\%$ real data outperforms models trained on all real data.
\end{abstract}

\vspace{-0.5cm}
\section{Introduction}

Imagine using an object detection system for an environment like your kitchen. Such a system needs to not only recognize different kinds of objects but also distinguish between many different \emph{instances} of the same object category, \eg, \emph{your} cup vs. \emph{my} cup. With the tremendous progress that has been made in visual recognition, as documented on benchmark detection datasets, one may expect to easily take a state-of-the-art system and deploy it for such a setting.

\begin{figure}[!t]
\centering
\includegraphics[width=0.48\textwidth]{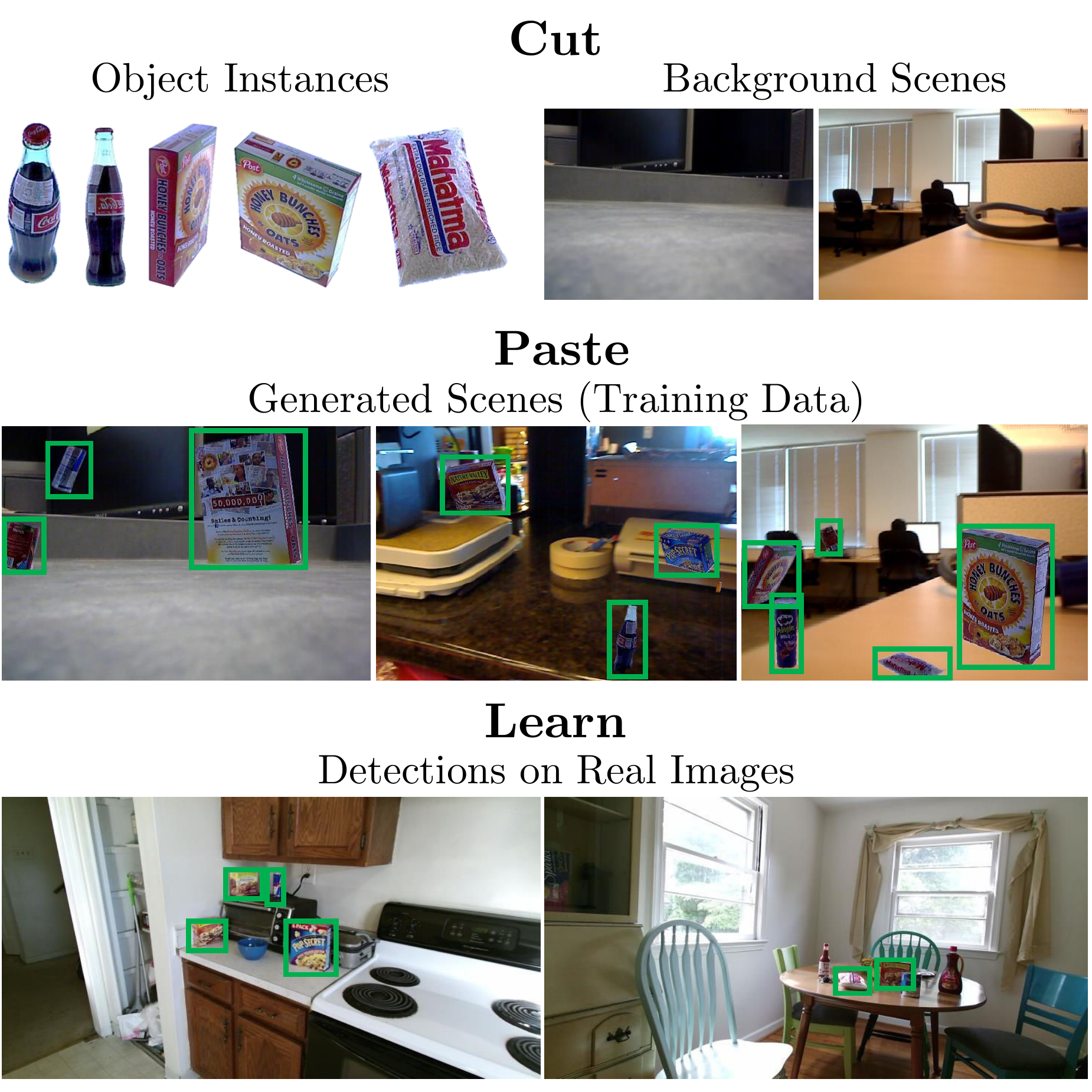}
\caption{We present a simple way to rapidly generate training images for instance detection with minimal human effort. We automatically extract object instance masks and render it on random background images to create realistic training images with bounding box labels. Our results show that such data is competitive with human curated datasets, and contains complementary information.}
\label{fig:teaser}
\end{figure}

\begin{figure*}[!t]
\centering
\includegraphics[width=0.98\textwidth]{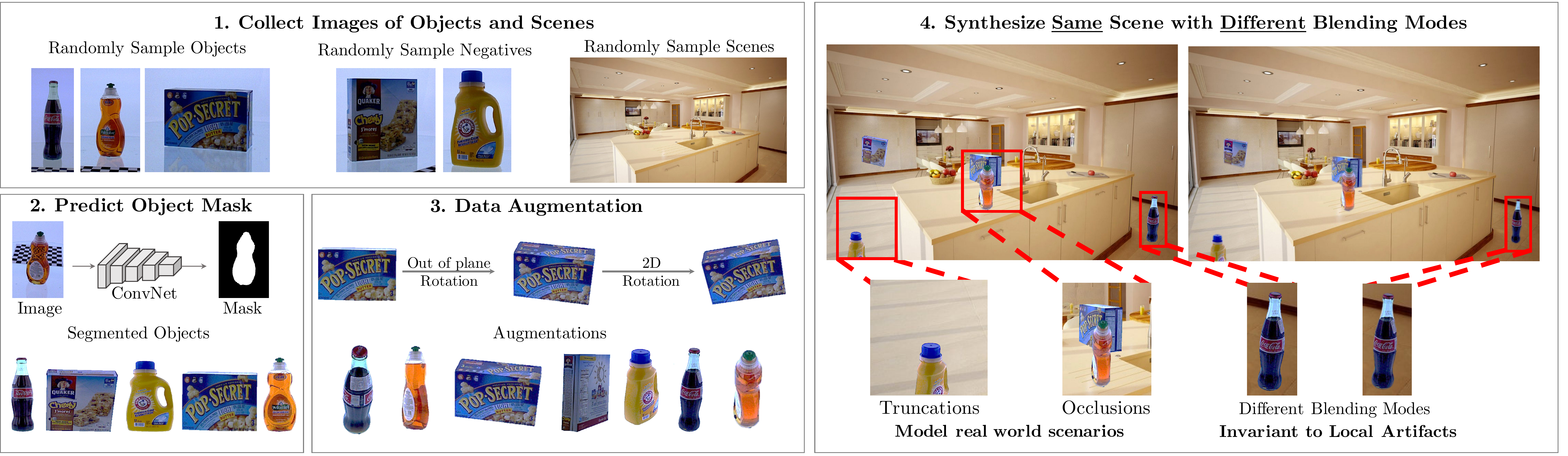}
\vspace{-0.1in}
\caption{We present a simple approach to rapidly synthesize datasets for instance detection. We start with a set of images of the instances and background scenes. We then automatically extract the object mask and segment the object. We paste the objects on the scenes with different blending to ensure that local artifacts are ignored by the detection model. Our results show that this synthesized data is both competitive with real data and contains complementary information.}
\vspace{-0.15in}
\label{fig:approach}
\end{figure*}

However, one of the biggest drawbacks of using a state-of-the-art detection system is the amount of annotations needed to train it. For a new environment with new objects, we would likely need to curate thousands of diverse images with varied backgrounds and viewpoints, and annotate them with boxes. Traditionally, vision researchers have undertaken such a mammoth task~\cite{PASCAL,COCO} for a few commonly occurring categories like \texttt{man}, \texttt{cow}, \texttt{sheep} \etc, but this approach is unlikely to scale to all possible categories, especially the instances in your kitchen.
In a personalized setting we need annotations for \emph{instances} like \emph{your} cup. We believe that collecting such annotations is a major impediment for rapid deployment of detection systems in robotics or other personalized applications.

Recently, a successful research direction to overcome this annotation barrier, is to use synthetically rendered scenes and objects~\cite{su2015render,movshovitz2016useful,karsch2011rendering} to train a detection system. This approach requires a lot of effort to make the scenes and objects realistic, ensuring high quality \emph{global and local consistency}. Moreover, models trained on such synthetic data have trouble generalizing to real data because of the change in image statistics~\cite{chen2016synthesizing,peng2015learning}. To address this, an emerging theme of work~\cite{gupta2016synthetic} moves away from graphics based renderings to composing real images. The underlying theme is to `paste' real object masks in real images, thus reducing the dependence on graphics renderings. 
Concurrent work~\cite{georgakis2017synthesizing} estimates scene geometry and layout and then synthetically places object masks in the scene to create realistic training images. However, the scene layout estimation step may not generalize to unseen scenes. In our paper, we show a simpler approach that does not require such scene geometry estimation to create training images.

Our key insight is that state-of-the art detection methods like Faster-RCNN~\cite{ren2015faster} and even older approaches like DPM~\cite{felzenszwalb2008discriminatively} \etc care more about \emph{local} region-based features for detection than the \emph{global} scene layout. As an example, a cup detector mostly cares about the visual appearance of the cup and its blending with the background, and not so much about where the cup occurs in the scene: the table-top or the ground. We believe that while global consistency is important, only ensuring \textit{patch-level realism} while composing synthetic datasets should go a long way to train these detectors. We use the term \textit{patch-level realism} to refer to the observation that the bounding box containing the pasted object \textit{looks} realistic to the human eye. 

However, naively placing object masks in scenes creates subtle pixel artifacts in the images. As these minor imperfections in the pixel space feed forward deeper into the layers of a ConvNet~\cite{lecun1989backpropagation}, they lead to noticeably different features and the training algorithm focuses on these discrepancies to detect objects, often ignoring to model their complex visual appearance. As our results show (Table ~\ref{tab:ablation}), such models give reduced detection performance.

Since our main goal is to create training data that is useful for training detectors, we resolve these local imperfections and maintain patch level realism. Inspired from methods in data augmentation and denoising auto encoders~\cite{vincent2008extracting}, we generate data that forces the training algorithm to ignore these artifacts and focus only on the object appearance. We show how rendering the same scene with the same object placement and only varying the blending parameter settings (Section~\ref{sec:adding-objs}) makes the detector robust to these subtle pixel artifacts and improves training. Although these images do not respect global consistency or even obey scene factors such as lighting \etc, training on them leads to high performance detectors with little effort. Our method is also complementary to existing work~\cite{movshovitz2016useful,su2015render,georgakis2017synthesizing} that ensures global consistency and can be combined with them.

Data generated using our approach is surprisingly effective at training detection models. Our results suggest that curated instance recognition datasets suffer from poor coverage of the visual appearances of the objects. With our method, we are able to generate many such images with different viewpoints/scales, and get a good coverage of the visual appearance of the object with minimal effort. Thus, our performance gain is particularly noticeable when the test scenes are different from the training scenes, and thus the objects occur in different viewpoints/scales.

\section{Related Work}
\vspace{-0.1in}
Instance detection is a well studied problem in computer vision.~\cite{zheng2016sift} provides a comprehensive overview of the popular methods in this field. Early approaches, such as ~\cite{collet2011moped}, heavily depend on extracting local features such as SIFT~\cite{lowe2004distinctive}, SURF~\cite{bay2008speeded}, MSER~\cite{mikolajczyk2005comparison} and matching them to retrieve instances~\cite{lowe2001local,tolias2015particular}. These approaches do not work well for objects which are not `feature-rich', where shape-based methods~\cite{hsiao2010making,ferrari2006object,hinterstoisser2012gradient} are more successful.

Modern detection methods~\cite{girshick2015fast,ren2015faster,girshick2014rich} based on learned ConvNet features~\cite{krizhevsky2012imagenet,simonyan2014very,lecun1989backpropagation} generalize across feature rich and feature poor objects~\cite{sharif2014cnn}. With the availability of powerful commodity GPUs, and fast detection algorithms~\cite{liu2016ssd,Redmon_2016_CVPR}, these methods are suitable for real-time object detection required in robotics. More recently, deep learning based approaches in computer vision are being adopted for the task of pose estimation of specific objects\cite{mitash2017self,zeng2016multi,wong2017segicp}. Improving instance detection and pose estimation in warehouses will be signifcantly useful for the perception pipeline in systems trying to solve the Amazon Picking Challenge\cite{correll2016lessons}.

The use of these powerful methods for object and instance detection requires large amounts of annotated data. This requirement is both impractical and expensive for rapidly deploying detection systems. Sythesizing data is one way to address this issue. ~\cite{su2015render,peng2015learning} use rendered images of objects to do both object detection and pose estimation. They render 3D models of objects from different viewpoints and place them against randomly sampled backgrounds.~\cite{movshovitz2016useful} also highlight the importance of using photo-realsitic models in training CNNs. 

There is a wide spectrum of work where rendered datasets are used for computer vision tasks. At one end, we have datasets with images of single objects on random backgrounds~\cite{su2015render,peng2015learning,movshovitz2016useful,park2015articulated}. On the other end, there are datasets where the entire scene is rendered~\cite{gaidon2016virtual,richter2016playing,handa2015scenenet}. On that spectrum our work lies in between as we do not render the whole world but use real images of both objects and backgrounds to compose new scenes. In this sense, our work closely related to contemporary work from~\cite{gupta2016synthetic} which generates synthetic data for localizing text in scenes.

Sedaghat \etal~\cite{sedaghat2015unsupervised} show how an annotated dataset can be created for the task of object pose estimation by taking videos by walking around the object.~\cite{held2016robust} uses synthetic data from~\cite{chang2015shapenet} for multi-view instance recognition.~\cite{massa2016deep} use real and synthetic images for 2D-3D alignment.

Similarly,~\cite{varol2017learning,chen2016synthesizing} render 3D humans in scenes and use this data for pose estimation. Tasks requiring dense annotation, such as segmentation, tracking \etc have also shown to benefit by using such approaches~\cite{handa2015scenenet,gaidon2016virtual,Ros_2016_CVPR,richter2016playing}.~\cite{song2015robot} shows a novel approach for collecting data of objects in a closed domain setting.~\cite{lai2011large,georgakis2016multiview,ammirato2017dataset} annotate 3D points belonging to an object in the point cloud reconstruction of a scene and propagate the label to all frames where the object is visible. As synthetic data can be significantly different from real images,~\cite{tzeng2015simultaneous} shows a domain adaptation approach to overcome this issue. In contrast, our work composes training scenes using real object images as well as real background images.

The existing approaches to sythesizing datasets focus largely on ensuring global consistency and realism~\cite{karsch2011rendering,varol2017learning,chen2016synthesizing,georgakis2017synthesizing}. While global consistency is important, we believe that local features matter more for training detection systems. Our approach ensures that when we train our detection model it is invariant to local discrepancies.
\begin{figure}[!t]
\centering
\includegraphics[width=0.49\textwidth]{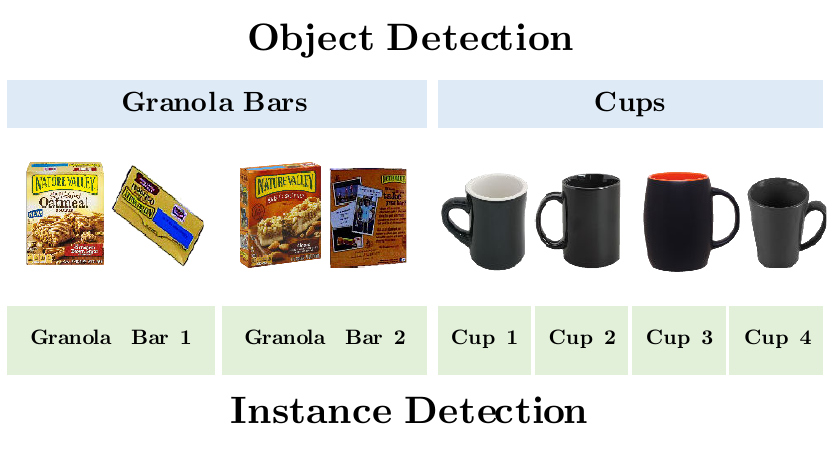}
\caption{Object vs Instance Detection. Instance detection involves fine-grained recognition within the same `object category'(as shown by the visually similar cups) while also detecting the same instance from different viewpoints(depicted by the different views of the granola bars). In this example, instance recognition must distinguish amongst 6 classes: 2 types of granola bars and 4 types of coffee cups. Object detection would distinguish only amongst 2 classes: coffee cups and granola bars.}

\label{fig:objvsinstancedetection}
\vspace{-1em}
\end{figure}

\section{Background}
\par \noindent \textbf{Instance Detection:} Instance detection requires accurate localization of a particular object, \eg a particular brand of cereal, a particular cup \etc. In contrast, generic object detection detects an entire generic category like a cereal box or a cup (see Figure~\ref{fig:objvsinstancedetection}). In fact, in the instance detection scenario correctly localizing a cereal box of some other brand is counted as a mistake.
Instance detection occurs commonly in robotics, AR/VR \etc, and can also be viewed as fine-grained recognition.

\par \noindent \textbf{Traditional Dataset Collection:} Building detection datasets involves a data curation step and an annotation step. Typically, data curation involves collecting internet images for object detection datasets~\cite{PASCAL,COCO}. However, this fails for instance datasets as finding internet images of particular instances is not easy. For instance detection~\cite{singh2014bigbird} data curation involves placing the instances in varied backgrounds and manually collecting the images. Manually collecting these images requires one to pay attention to ensure diversity in images by placing the object in different backgrounds and collecting different viewpoints.
The annotation step is generally crowd sourced. Depending on the type of data, human annotations can be augmented with object tracking or 3D sensor information~\cite{lai2011large,georgakis2016multiview,ammirato2017dataset,song2015robot,vondrick2013efficiently}. 

Unfortunately, both these steps are not suitable for \emph{rapidly} gathering instance annotations. Firstly, as we show in our experiments, even if we limit ourselves to the same \emph{type} of scene, \eg, kitchens, the curation step can lack diversity and create biases that do not hold in the test setting. Secondly, as the number of images and instances increase, manual annotation requires additional time and expense.
\section{Approach Overview}

We propose a simple approach to rapidly collect data for instance detection. Our results show that our approach is competitive with the manual curation process, while requiring little time and no human annotation. 

Ideally, we want to capture all of the visual diversity of an instance. Figures~\ref{fig:teaser} and~\ref{fig:objvsinstancedetection} show how a single instance appears different when seen from different views, scales, orientation and lighting conditions. Thus, distinguishing between such instances requires the dataset to have good coverage of viewpoints and scales of the object. Also, as the number of classes increases rapidly with newer instances, the long-tail distribution of data affects instance recognition problems. With synthetic data, we can ensure that the data has good coverage of both instances and viewpoints. Figure~\ref{fig:approach} shows the main steps of our method:

\begin{enumerate}[noitemsep,leftmargin=*]
    \item \textbf{Collect object instance images:} Our approach is agnostic to the way the data is collected. We assume that we have access to object images which cover diverse viewpoints and have a modest background.
    \item \textbf{Collect scene images:} These images will serve as background images in our training dataset. If the test scenes are known beforehand (like in the case of a smart-home or a warehouse) one can collect images from those scenes. As we do not compute any scene statistics like geometry or layout, our approach can readily deal with new scenes.
    \item \textbf{Predict foreground mask for the object:} We predict a foreground mask which separates the instance pixels from the background pixels. This gives us the object mask which can be placed in the scenes.
    \item \textbf{Paste object instances in scenes:} Paste the extracted objects on a randomly chosen background image. We ensure \textbf{invariance to local artifacts} while placing the objects so that the training algorithm does not focus on subpixel discrepancies at the boundaries. We add various modes of blending and synthesize the exact same scene with different blending to make the algorithm robust to these artifacts. We also add \textbf{data augmentation} to ensure a diverse viewpoint/scale coverage.
\end{enumerate}

\vspace{-0.5em}
\section{Approach Details and Analysis}
\label{sec:approach-detail}

\begin{figure*}[t]
\centering
\includegraphics[width=0.98\textwidth]{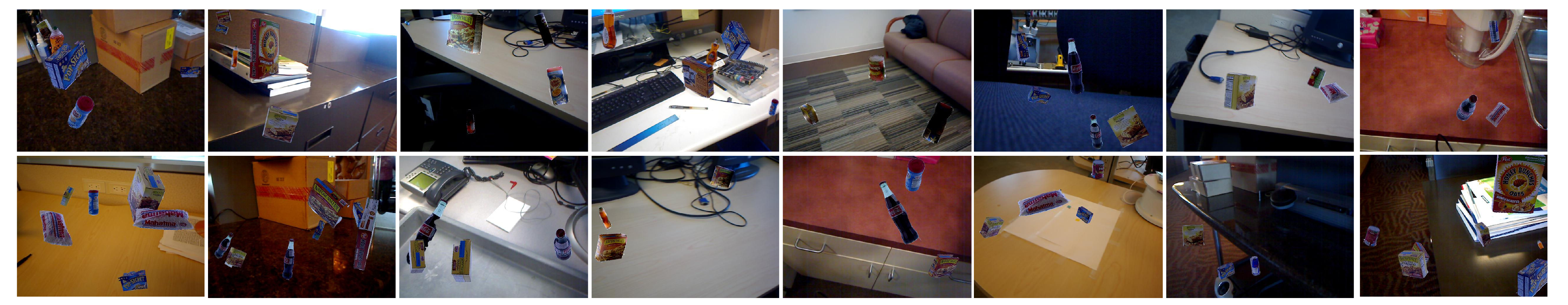}
\caption{A few randomly chosen samples from our synthesized images. We describe the details of our approach in Section~\ref{sec:approach-detail}.}
\label{fig:genimgs}
\end{figure*}

\begin{table*}[!t]
\setlength{\tabcolsep}{0.3em}
\centering
\footnotesize{
\caption{We analyze the effect of various factors in synthesizing data by generating data with different settings and training a detector~\cite{ren2015faster}. We evaluate the trained model on the GMU Dataset~\cite{georgakis2016multiview}. As we describe in Section~\ref{sec:approach-detail}, these factors greatly improve the quality of the synthesized data.}
\label{tab:ablation}
    \begin{tabular}{@{}lccccccccccccccccc@{}}
        \toprule
         & 2D Rot. & 3D Rot. & Trunc.& Occl.  & coca & coffee & honey 
        & hunt's & mahatma & nature & nature 
        & palmolive & pop & pringles & red
        & mAP\\
        
        & & &  & &  cola & mate & bunches 
        & sauce & rice & v1 & v2 
        & orange & secret & bbq & bull
        &\\
        
        \midrule
        \textbf{Blending} (Sec~\ref{sec:blending}) & &  & & & &  &  &  &  &  & & & & & & \\
        No blending & \checkmark & \checkmark & \checkmark & \checkmark & 65.7 & 91.1 & \textbf{83.2} & 59.8 & 57.7 & 92.1 & 84.4 & 61.4 & 59.0 & 38.7 & 31.9 & 65.9\\
        Gaussian Blurring & \checkmark & \checkmark & \checkmark & \checkmark & 65.3 & 88.1 & 80.8 & \textbf{67.5} & 63.7 & 90.8 & 79.4 & 57.9 & 58.9 & \textbf{65.7} & 40.2 & 68.9\\
         Poisson~\cite{perez2003poisson} & \checkmark & \checkmark & \checkmark & \checkmark & 62.9 & 82.9 & 63.9 & 59.4 & 20.7 & 84.6 & 67.9 & 60.9 & 73.5 & 41.0 & 25.1 & 58.4\\
        All Blend & \checkmark & \checkmark & \checkmark & \checkmark & 76.0 & 90.3 & 79.9 & 65.4 & 67.3 & \textbf{93.4} & \textbf{86.6} & 64.5 & 73.2 & 60.4 & 39.8 & 72.4 \\        
        All Blend + same image & \checkmark & \checkmark & \checkmark & \checkmark & \textbf{78.4} & \textbf{92.7} & 81.8 & 66.2 & \textbf{69.8} & 93.0 & 82.9 & \textbf{65.7} & \textbf{76.0} & 62.9 & \textbf{41.2} & \textbf{73.7} \\        
        \midrule
        \textbf{Data Aug.} (Sec~\ref{sec:data-aug}) &  & & & & &  &  &  &  &  & & & & & & \\        
        No 2D Rotation &  & \checkmark & \checkmark & \checkmark & 63.3 & 90.4 & 81.4 & 63.7 & 54.2 & 91.8 & 82.3 & 59.2 & 71.3 & 68.2 & 41.4 & 69.7 \\
        No 3D Rotation & \checkmark  & & \checkmark & \checkmark & 73.1 & 90.6 & 83.2 & 63.3 & 55.8 & 93.4 & 82.1 & 65.9 & 64.5 & 45.7 & 33.6 & 68.3\\
        No Trunc. & \checkmark  & \checkmark &  & \checkmark & 73.4 & 92.1 & 77.8 & 59.9 & 64.6 & 92.4 & 84.6 & 62.0 & 74.2 & 67.4 & 41.7 & 71.8 \\
        No Occlusion & \checkmark  & \checkmark & \checkmark   & & 63.1 & 84.9 & 74.4 & 64.5 & 50.8 & 76.9 & 67.6 & 55.7 & 69.0 & 58.7 & 28.1 & 63.1\\
        All & \checkmark & \checkmark & \checkmark & \checkmark & 78.4 & 92.7 & 81.8 & \bfline{66.2} & 69.8 & 93.0 & 82.9 & 65.7 & 76.0 & 62.9 & 41.2 & 73.7\\
        All + Distractor & \checkmark  & \checkmark & \checkmark & \checkmark  & \bfline{81.0} & \bfline{93.3} & \bfline{85.6} & 55.6 & \bfline{73.8} & \bfline{94.9} & \bfline{87.1} & \bfline{68.7} & \bfline{79.5} & \bfline{77.1} & \bfline{42.0} & \bfline{76.2}\\
        \bottomrule
    \end{tabular}
}
\end{table*}

We now present additional details of our approach and provide empirical analysis of our design choices.

\subsection{Collecting images}
We first describe how we collect object/background images, and extract object masks without human effort.

\par \noindent \textbf{Images of objects from different viewpoints:} We choose the objects present in Big Berkeley Instance Recognition Dataset (BigBIRD)~\cite{singh2014bigbird} to conduct our experiments. Each object has 600 images, captured by five cameras with different viewpoints. Each image also has a corresponding depth image captured by an IR camera.

\par \noindent \textbf{Background images of indoor scenes:} We place the extracted objects from the BigBIRD images on randomly sampled background images from the UW Scenes dataset~\cite{lai2011large}. There are 1548 images in the backgrounds dataset.

\label{sec:fgbg}
\par
\begin{figure}[!t]
\centering
\includegraphics[width=0.48\textwidth]{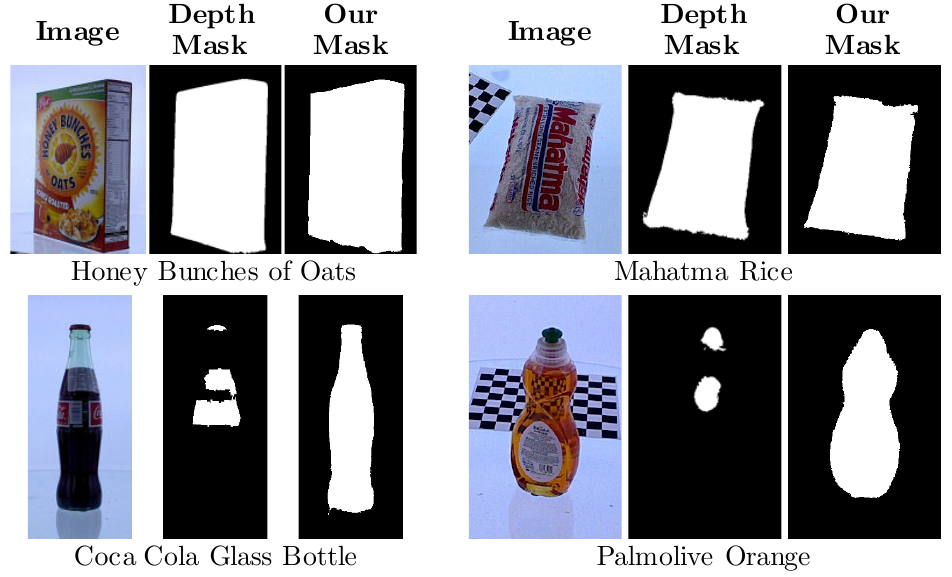}
\caption{Given an image of a new unseen object instance, we use a ConvNet to predict foreground/background pixels. Using these predictions we automatically obtain an object mask. This method generalizes to transparent surfaces where traditional methods relying on depth sensors for segmentation fail (second row).}
\label{fig:fgbg}
\vspace{-1em}
\end{figure}

 \noindent \textbf{Foreground/Background segmentation:} Once we have collected images of the instances, we need to determine the pixels that belong to the instance vs. the background. We automate this by training a model for foreground/background classification.
We train a FCN network~\cite{long2015fully} (based on VGG-16~\cite{simonyan2014very} pre-trained on PASCAL VOC~\cite{PASCAL} image segmentation) to classify each image pixel into foreground/background. The object masks from the depth sensor are used as ground truth for training this model. We train this model using images of instances which are not present in our final detection evaluation.
We use~\cite{barron2016fast} as a post-processing step to clean these results and obtain an object mask. Figure~\ref{fig:fgbg} shows some of these results. In practice, we found this combination to generalize to images of unseen objects with modest backgrounds and give good quality object masks from input images. It also generalizes to transparent objects, \eg, \texttt{coca cola} bottle, where the depth sensor does not work well.

\vspace{-0.5em}

\subsection{Adding Objects to Images}
\label{sec:adding-objs}
After automatically extracting the object masks from input images, we paste them on real background images. Na{\"i}vely pasting objects on scenes results in artifacts which the training algorithm focuses on, ignoring the object's visual appearance. In this section, we present steps to generate data that forces the training algorithm to ignore these artifacts and focus only on the object appearance.
To evaluate these steps empirically, we train a detection model on our synthesized images and evaluate it on a benchmark instance detection dataset (real images).

\par \noindent \textbf{Detection Model:} We use the Faster R-CNN~\cite{ren2015faster} method and initialize the model from a VGG-16~\cite{simonyan2014very} model pre-trained on object detection on the MSCOCO~\cite{COCO} dataset.

\par \noindent \textbf{Benchmarking Dataset:} After training the detection model on our synthetic images, we use the GMU Kitchen dataset~\cite{georgakis2016multiview} for evaluation. There are 9 scenes in this dataset. Three dataset splits with 6 scenes for training and 3 for testing have been provided in ~\cite{georgakis2016multiview} to conduct experiments on the GMU Kitchen dataset. We follow these splits for train/test and report the average over them. No images or statistics from this dataset are used for either dataset synthesis or training the detector. We report Mean Average Precision (mAP) at IOU of 0.5~\cite{PASCAL} in all our experiments.

\vspace{-1em}
\subsubsection{Blending}
\label{sec:blending}
\begin{figure}[!t]
\centering
\includegraphics[width=0.48\textwidth]{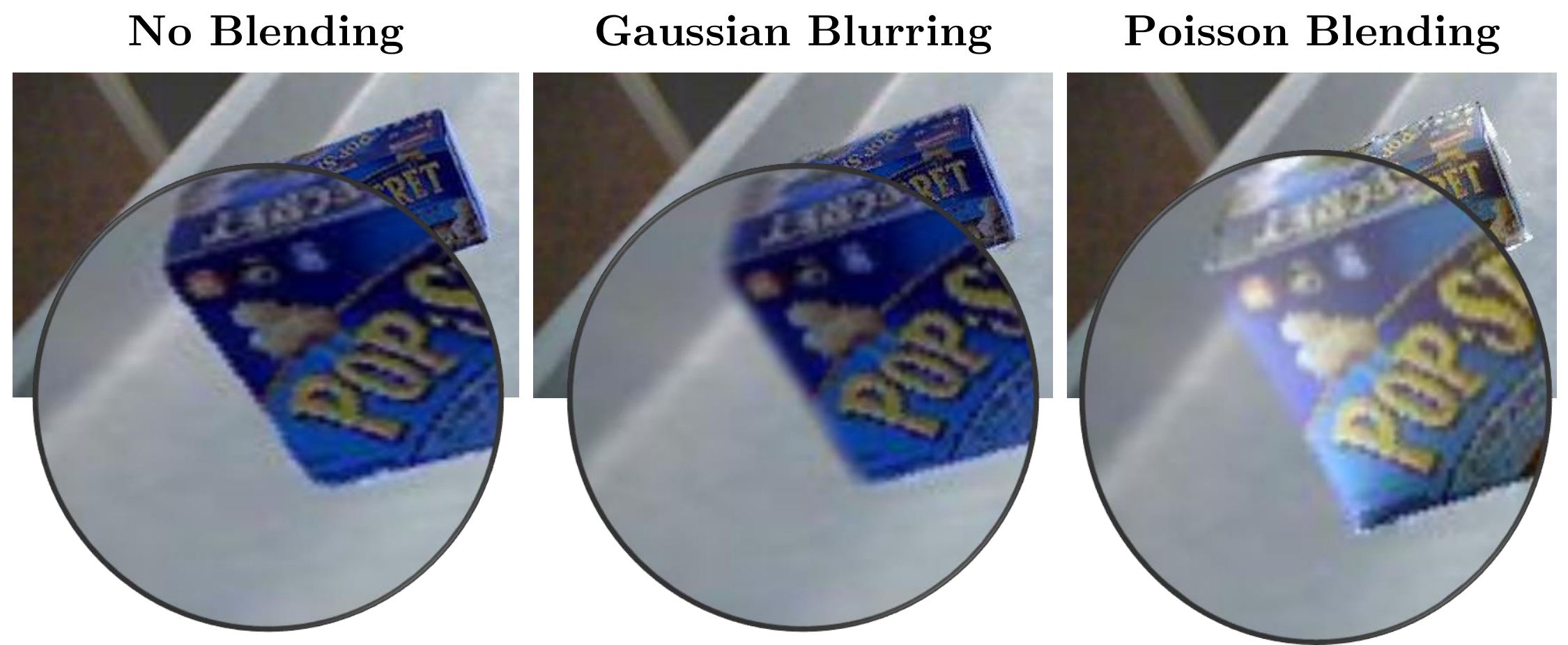}
\caption{Different blending modes used while generating datasets. These modes help the model in ignoring artifacts arising from pasting objects on background scenes. More details in Section~\ref{sec:blending}}
\label{fig:blendingmodes}
\end{figure}

Directly pasting objects on background images creates boundary artifacts. Figure~\ref{fig:blendingmodes} shows some examples of such artifacts. Although these artifacts seem subtle, when such images are used to train detection algorithms, they give poor performance as seen in Table~\ref{tab:ablation}.
As current detection methods~\cite{ren2015faster} strongly depend on local region-based features, boundary artifacts substantially degrade their performance. 

The blending step `smoothens' out the boundary artifacts between the pasted object and the background. Figure~\ref{fig:blendingmodes} shows some examples of blending. Each of these modes add different image variations, \eg, Poisson blending~\cite{perez2003poisson} smooths edges and adds lighting variations. Although these blending methods do not yield visually `perfect' results, they improve performance of the trained detectors. Table~\ref{tab:ablation} lists these blending methods and shows the improvement in performance after training on blended images.

To make the training algorithm further ignore the effects of blending, we synthesize the exact same scene with the same object placement, and only vary the type of blending used. We denote this by `All Blend + same image' in Table~\ref{tab:ablation}. Training on multiple such images where only the blending factor changes makes the training algorithm invariant to these blending factors and \textbf{improves performance by 8 AP points} over not using any form of blending.

\subsubsection{Data Augmentation}
While pasting the objects on background, we also add the following modes of data augmentation:

\label{sec:data-aug}
\par \noindent \textbf{2D Rotation:} The objects are rotated at uniformly sampled random angles in between $30$ to $-30$ degrees to account for camera/object rotation changes. Table~\ref{tab:ablation} shows a gain of 3 AP points by adding this augmentation. 

\par
\begin{figure}[!t]
\centering
\includegraphics[width=0.48\textwidth]{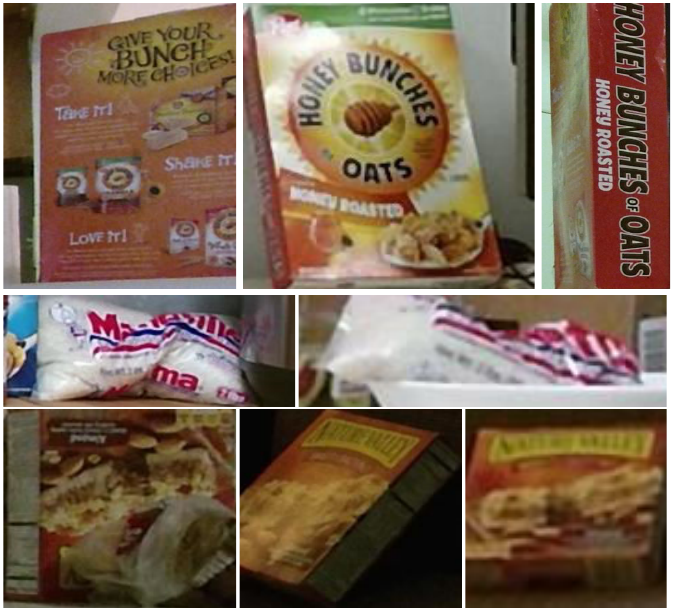}
\caption{Missed detections on the Active Vision Dataset~\cite{ammirato2017dataset} for a model trained on the hand-annotated GMU Dataset~\cite{georgakis2016multiview}. The model consistently fails to detect certain viewpoints as the training data has poor viewpoint coverage and has biases different from the test set. Each row shows a single instance.}
\label{fig:easymiss}
\end{figure}

 \noindent \textbf{3D Rotation:} As we can control this step, we have many images containing atypical 3D rotations of the instances which is hard to find in real data.
Table~\ref{tab:ablation} shows a gain of more than 4 AP points because of this augmentation. In Section~\ref{sec:exp-eval-active} and Figure~\ref{fig:easymiss}, we show examples of how a model trained on human collected data consistently fails to detect instances from certain viewpoints because the training data has poor viewpoint coverage and different biases from the test set. This result shows the value of being able to synthesize data with diverse viewpoints.

\par \noindent \textbf{Occlusion and Truncation:} Occlusion and truncation naturally appear in images. They refer to partially visible objects (such as those in Figure~\ref{fig:approach}). We place objects at the boundaries of the images to model truncation, ensuring at least 0.25 of the object box is in the image. To add occlusion, we paste the objects with partial overlap with each other (max IOU of $0.75$). Like other modes of augmentation, we can easily vary the amount of truncation/occlusion. As Table~\ref{tab:ablation} shows, adding truncation/occlusion improves the result by as much as 10 AP points. 

\par \noindent \textbf{Distractor Objects:} We add distractor objects in the scenes. This models real-world scenarios with multiple distractor objects. We use additional objects from the BigBIRD dataset as distractors. Presence of synthetic distractors also encourages the learning algorithm to not only latch on to boundary artifacts when detecting objects but also improves performance by 3 AP points.

\begin{figure}[!t]
\centering
\includegraphics[width=0.48\textwidth]{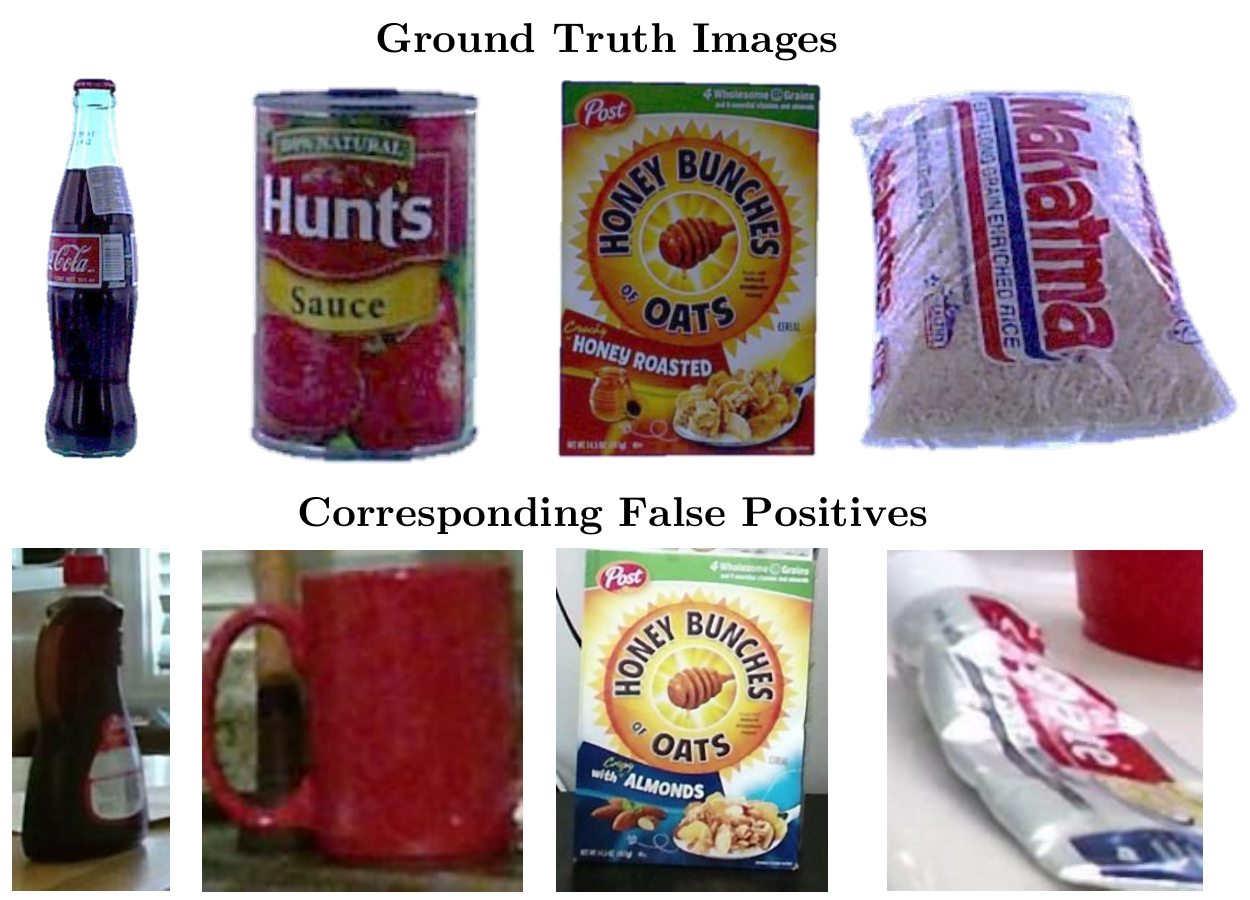}
\caption{Examples of false positives from the UNC dataset by the detector trained on the hand-annotated bounding boxes from the GMU dataset. Object detectors trained on hand annotated scenes also need new negatives to be able to perform well in newer scenes. }
\label{fig:fp}
\end{figure}
\begin{table*}[t]
\setlength{\tabcolsep}{0.3em}
\centering
\footnotesize{
\caption{We compute the performance of training a model on synthetic data and compare it against training on real data. We evaluate on the test split of the GMU Kitchen Dataset~\cite{georgakis2016multiview}.}
\label{table:realsyngmu}
    \begin{tabular}{@{}lcccccccccccc@{}}
    \toprule
        Dataset & coca & coffee & honey
        & hunt's & mahatma & nature & nature 
        & palmolive & pop & pringles & red
        & mAP\\
         & cola & mate & bunches 
        & sauce & rice & v1 & v2 
        & orange & secret & bbq & bull
        & \\
        \midrule
          Real Images from GMU  & 81.9 & 95.3 & 92.0 & 87.3 & 86.5 & 96.8 & 88.9 & \textbf{80.5} & 92.3 & 88.9 & 58.6 & 86.3\\
          \midrule
          SP-BL-SS~\cite{georgakis2017synthesizing} & 55.5 & 67.9 & 71.2 & 34.6 & 30.6 & 82.9 & 66.2 & 33.1 & 54.3 & 54.8 & 17.7 & 51.7\\
          (Ours) Synthetic Images &81.0 & 93.3 & 85.6 & 55.6 & 73.8 & 94.9 & 87.1 & 68.7 & 79.5 & 77.1 & 42.0 & 76.2\\
          \midrule
          SP-BL-SS + Real Images~\cite{georgakis2017synthesizing}  & 82.6 & 92.9 & 91.4 & 85.5 & 81.9 & 95.5 & 88.6 & 78.5 & 93.6 & 90.2 & 54.1 & 85.0 \\
          (Ours) Synthetic + Real Images & \textbf{88.5} & \textbf{95.5} & \textbf{94.1} & \textbf{88.1} & \textbf{90.3} & \textbf{97.2} & \textbf{91.8} & 80.1 & \textbf{94.0} & \textbf{92.2} &\textbf{ 65.4 }& \textbf{88.8}\\          
          \bottomrule
    \end{tabular}
}    
\end{table*}

\section{Experiments}

We now compare the effectiveness of our synthesized data against human annotated data on two benchmark datasets. We first describe our common experimental setup.
 
\par \noindent \textbf{Synthesized Data:}
We analyze our design choices in Section~\ref{sec:approach-detail} to pick the best performing ones. We use a total of 33 object instances from the BigBIRD Dataset~\cite{singh2014bigbird} overlapping with the 11 instances from GMU Kitchen Dataset~\cite{georgakis2016multiview} and the 33 instances from Active Vision Dataset~\cite{ammirato2017dataset}.
We use a foreground/background ConvNet (Section~\ref{sec:fgbg}) to extract the foreground masks from the images. The foreground/background ConvNet is \emph{not} trained on instances we use to evaluate detection. As in Section~\ref{sec:approach-detail}, we use backgrounds from the UW Scenes Dataset~\cite{lai2011large}
We generate a synthetic dataset with approximately 6000 images using all modes of data augmentation from Section~\ref{sec:approach-detail}. We sample scale, rotation, position and the background randomly. Each background appears roughly 4 times in the generated dataset with different objects. To model occlusions we allow a maximum IOU of 0.75 between objects. For truncations, we allow at least 25\% of the object box to be in the image. For each scene we have three versions produced with different blending modes as described in Section~\ref{sec:blending}. Figure~\ref{fig:genimgs} shows samples of generated images. We use this synthetic data for all our experiments. The code used for generating scenes is available at: \url{https://goo.gl/imXRt7}.

\par \noindent \textbf{Model:} We use a Faster R-CNN model~\cite{ren2015faster} based on the VGG-16~\cite{simonyan2014very} pre-trained weights on the MSCOCO~\cite{COCO} detection task. We initialize both the RPN trunk and the object classifier trunk of the network in this way.
We fine-tune on different datasets (both real and synthetic) and evaluate the model's performance. We fine-tune all models for $25$K iterations using SGD+momentum with a learning rate of $0.001$, momentum $0.9$, and reduce the learning rate by a factor of $10$ after $15$K iterations. We also use weight decay of $0.0005$ and dropout of $0.5$ on the fully-connected layers. We set the value of all the loss weights (both RPN and classification) as $1.0$ in our experiments. We ensure that the model hyperparameters and random seed do not change across datasets/experiments for consistency.

\par \noindent \textbf{Evaluation:} We report Average Precision (AP) at IOU of $0.5$ in all our experiments for the task of instance localization. Following~\cite{ammirato2017dataset}, we consider boxes of size at least $50 \times 30$ pixels in the images for evaluation.

\begin{table}[!t]
\setlength{\tabcolsep}{0.17em}
\centering
\footnotesize{
\caption{Evaluation on the entire Active Vision dataset by varying the amount of real data from the GMU Kitchen Scenes \emph{train} dataset}
\label{table:realsynunc}
    \begin{tabular}{@{}lccccccc@{}}
        \toprule
        Dataset & coca & honey
        & hunt's & mahatma & nature & red
        & mAP\\ 
        & cola  & bunches 
        & sauce & rice & v2 & bull
        & \\ 
        \midrule        
        Real Images  & 57.7 & 34.4 & 48.0 & 39.9 & 24.6 & 46.6 & 41.9 \\
        Synthetic  & 63.0 & 29.3 & 34.2 & 20.5 & \textbf{49.0}  & 23.0 & 36.5\\          
        Synthetic + Real Images & \textbf{69.9} &\textbf{ 44.2} & \textbf{51.0} & \textbf{41.8} & 48.7 & \textbf{50.9} & \textbf{51.1}\\ 
        \midrule
        10\% Real   & 15.3 & 19.1 & 31.6 & 11.2 & 6.1 & 11.7 & 15.8 \\
        10\% Real + Syn   & 66.1 & 36.5 & 44.0 & 26.4 & 48.9 & 37.6 & 43.2\\
        40\% Real   & 55.8 & 31.6 & 47.3 & 27.4 & 24.8 & 41.9 & 38.2 \\
        40\% Real + Syn  & \textbf{69.8} & 41.0 & \textbf{55.7} & 38.3 & \textbf{52.8} & 47.0 & \textbf{50.8}\\ 
        70\% Real   & 55.3 & 30.6 & 47.9 & 36.4 & 25.0 & 41.2 & 39.4 \\
        70\% Real + Syn & 67.5 & \textbf{42.0} & 50.9 & \textbf{43.0} & 48.5 & \textbf{51.8} & 50.6\\ 
        \bottomrule
    \end{tabular}
    }
\end{table}

\subsection{Training and Evaluation on the GMU Dataset}
\label{sec:exp-gmu}

Similar to Section~\ref{sec:approach-detail}, we use the GMU Kitchen Dataset\cite{georgakis2016multiview} which contains $9$ kitchen scenes with $6,728$ images. We evaluate on the 11 objects present in the dataset overlapping with the BigBIRD~\cite{singh2014bigbird} objects. We additionally report results from~\cite{georgakis2017synthesizing}. Their method synthesizes images by accounting for global scene structure when placing objects in scenes, \eg, ensure that cups lie on flat surfaces like table tops. In contrast, our method does not use take into account such global structure, but focuses on patch-level realism. We note that their method~\cite{georgakis2017synthesizing} uses a different background scenes dataset for their synthesis.

\begin{figure*}[t]
\centering
\includegraphics[width=0.99\textwidth]{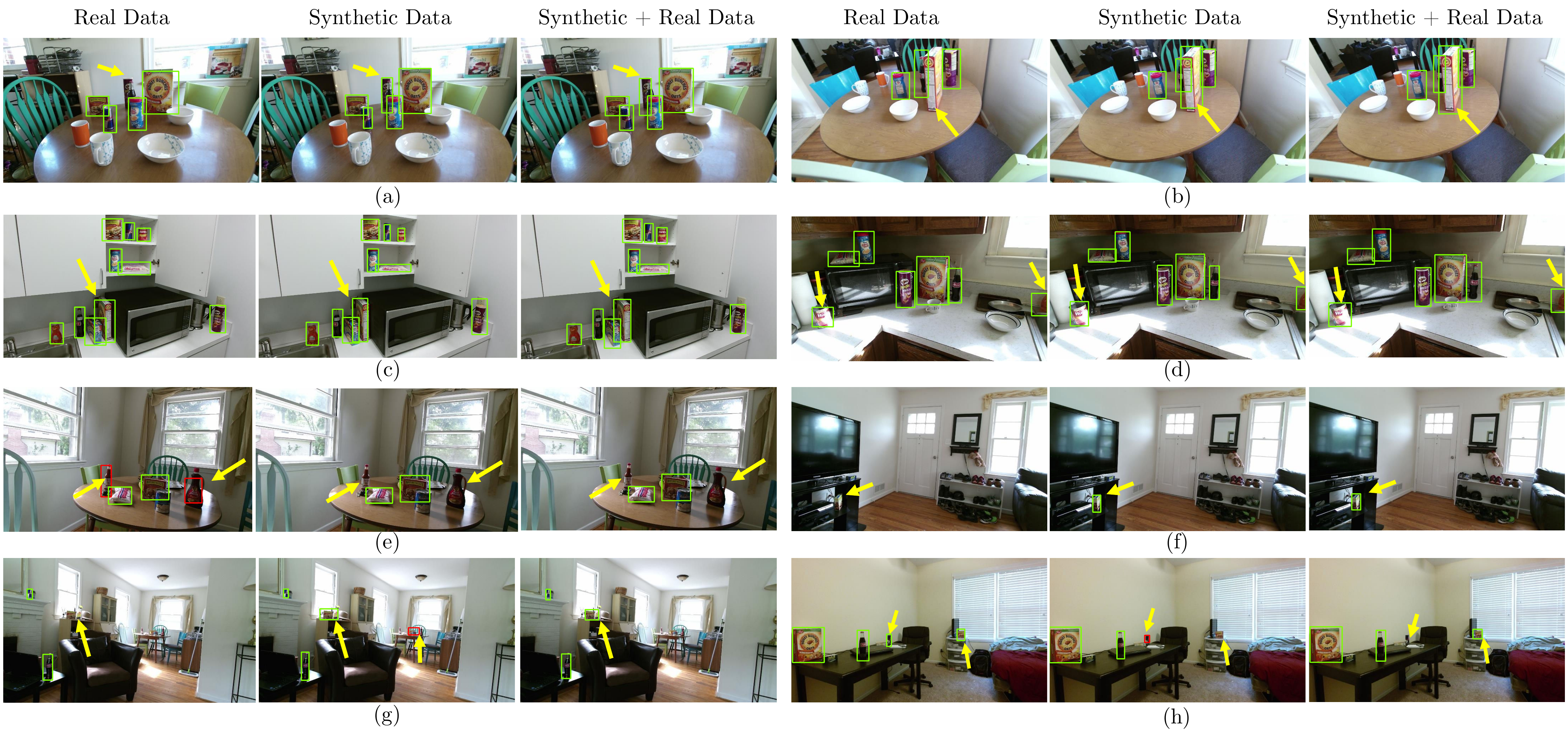}
\vspace{-0.15in}
\caption{We show qualitative detection results and mark true positives in green, false positives in red and arrows to highlight regions. The top two rows are from the GMU Kitchen Scenes~\cite{georgakis2016multiview} and the bottom two rows from the Active Vision Dataset~\cite{ammirato2017dataset}. $(a),(b)$: Model trained on real data misses objects which are heavily occluded $(a)$ or stops detecting objects as viewpoint changes from $a$ to $b$.
$(c),(d)$: Model trained on synthetic data detects occluded and truncated objects.
$(e)$: Combining synthetic data removes false positives due to training only on real data.
$(g),(h)$: Combining real data removes false positives due to training only on synthetic data. $(f),(g)$: Viewpoint changes cause false negatives. \footnotesize{(Best viewed electronically)}}

\vspace{-0.15in}
\label{fig:detections}
\end{figure*}

Table ~\ref{table:realsyngmu} shows the evaluation results. We see that training on the synthetic data is competitive with training on real images (rows 1 vs 3) and also outperforms the synthetic dataset from~\cite{georgakis2017synthesizing} (rows 2 vs 3). Combining synthetic data with the real data shows a further improvement for all synthetic image datasets (rows 4, 5). These results show that the data generated by our approach is not only competitive with both real data and existing synthetic data, but also provides complementary information. Figure~\ref{fig:detections} shows qualitative examples illustrating this point.

\subsection{Evaluation on the Active Vision Dataset}
\label{sec:exp-eval-active}
To test generalization across datasets, we now present experiments where we train on either our synthetic data or the GMU Dataset~\cite{georgakis2016multiview}, and evaluate on the Active Vision Dataset~\cite{ammirato2017dataset}. The Active Vision Dataset\cite{ammirato2017dataset} has 9 scenes and 17,556 images. It has 33 objects in total and 6 objects in overlap with the GMU Kitchen Scenes. We use these 6 objects for our analysis. We do \emph{not} use this dataset for training.

We train a model trained on \emph{all} the images from the GMU Dataset (Section~\ref{sec:exp-gmu}). This model serves as a baseline for our model trained on synthetic data. As Table~\ref{table:realsynunc} shows, by collecting just 10\% images and adding our synthetically generated images, we are able to get more MAP than using the real images in the dataset without the synthetic images. This highlights how useful our approach of dataset generation is in scenarios where there is a dearth of labeled images. Also, the performance gap between these datasets is smaller than in Section~\ref{sec:exp-gmu}.

\par \noindent \textbf{Failure modes of real data:} Upon inspecting the errors~\cite{hoiem2012diagnosing} made by the GMU model, we see that a common error mode of the detector is its \textbf{failure to recognize certain views} in the test-set (see Figure ~\ref{fig:easymiss}). These viewpoints were sparsely present in the human annotated training data. In contrast, our synthetic training data has a diverse viewpoint coverage. The model trained on the synthesized images drastically reduces these errors. Combining the synthesized images with the real images from GMU gives a further improvement of \textbf{10 AP points} suggesting that synthesized images do provide complementary information.

\par \noindent \textbf{Varying Real Data:} We investigate the effect of varying the number of real images combined with the synthesized data. We randomly sample different amounts of real images from the GMU Dataset and combine them with the synthetic data to train the detector. As a baseline we also train the model on varying fractions of the real data. Table~\ref{table:realsynunc} shows that by adding synthetic images to just $10\%$ of the real images we get a boost of \textbf{10 AP points} over just using real images. This performance is also tantalizingly close to the performance of combining larger fractions of real data. This result reinforces the effectiveness and complementary nature of our approach. In the supplementary material, we present additional such results.
\section{Discussion and Future Work}

We presented a simple technique to synthesize annotated training images for instance detection. Our key insights were to leverage randomization for blending objects into scenes and to ensure a diverse coverage of instance viewpoints and scales. We showed that patch-based realism is sufficient for training region-proposal based object detectors. Our method performs favorably to existing hand curated datasets and captures complementary information. In a realistic cross-domain setting we show that by combining just $10\%$ of the available real annotations with our synthesized data, our model performs better than using all the real annotations. From a practical standpoint our technique affords the possibility of generating scenes with non-uniform distributions over object viewpoints and scales without additional data collection effort.

We believe our work can be combined with existing approaches ~\cite{georgakis2017synthesizing} that focus on global consistency for placing objects and~\cite{karsch2011rendering} which model realism. Future work should focus on a combination of such approaches.

\footnotesize{\par \noindent \textbf{Acknowledgements}: The authors are grateful to Georgios Georgakis and Phil Ammirato for their help with the datasets and discussions. This work was supported in part by NSF Grant CNS1518865.}

{\small
\bibliographystyle{ieee}
\bibliography{myRefs}
}

\clearpage
 \onecolumn
\section*{Appendix}
\begin{figure}[!h]
\centering
\includegraphics[height=0.9\textheight]{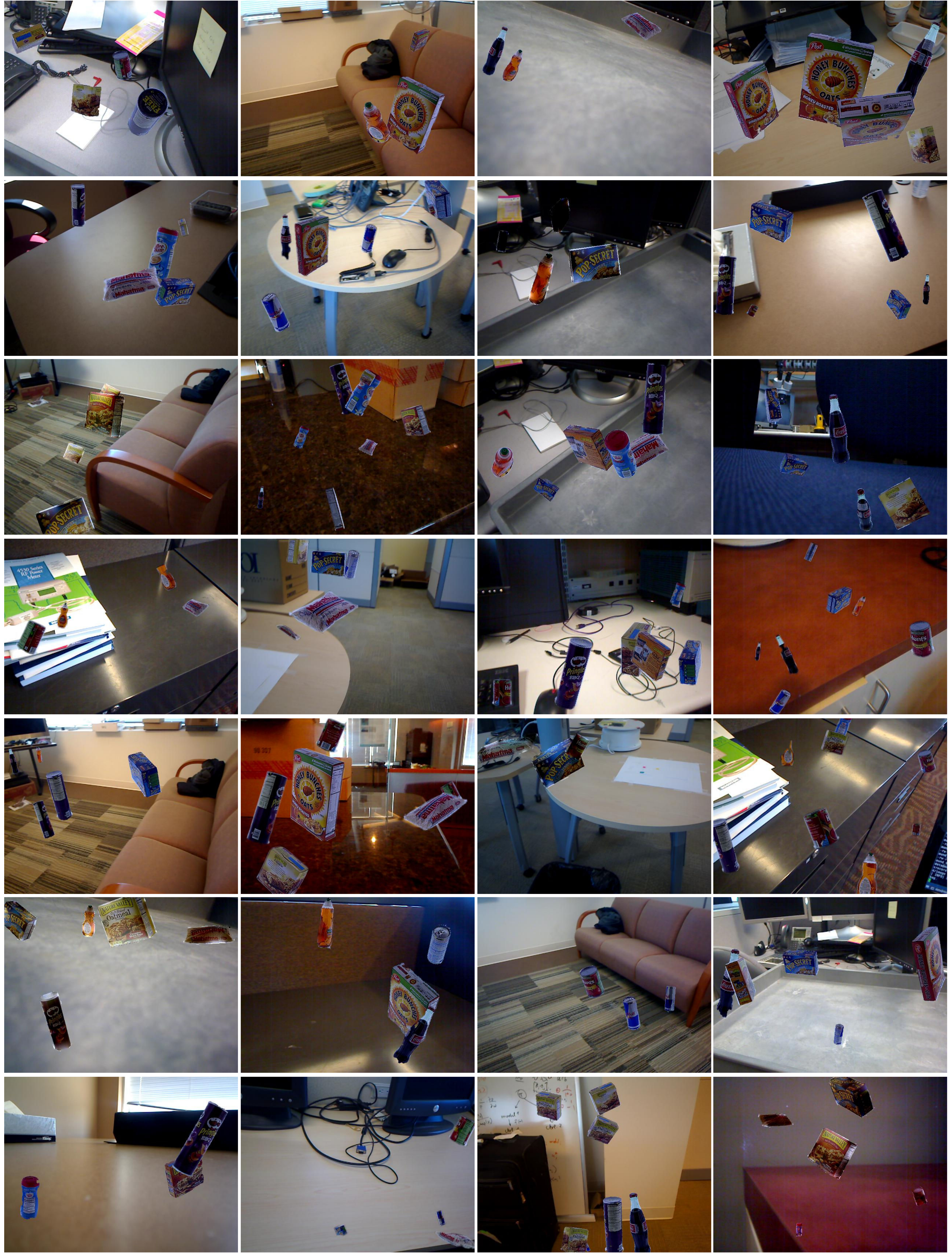}
\caption{A few randomly chosen samples from our synthesized images.}
\end{figure}

\end{document}